# The Cumulative Rule for Belief Fusion

Audun Jøsang

*QUT, Brisbane, Australia*

**Abstract**

The problem of combining beliefs in the Dempster-Shafer belief theory has attracted considerable attention over the last two decades. The classical Dempster's Rule has often been criticised, and many alternative rules for belief combination have been proposed in the literature. The consensus operator for combining beliefs has nice properties and produces more intuitive results than Dempster's rule, but has the limitation that it can only be applied to belief distribution functions on binary state spaces. In this paper we present a generalisation of the consensus operator that can be applied to Dirichlet belief functions on state spaces of arbitrary size. This rule, called the cumulative rule of belief combination, can be derived from classical statistical theory, and corresponds well with human intuition.

*Key words:* Belief theory, belief fusion, consensus operator

## 1 Introduction

Belief theory has its origin in a model for upper and lower probabilities proposed by Dempster in 1960. Shafer later proposed a model for expressing beliefs [1]. The main idea behind belief theory is to abandon the additivity principle of probability theory, i.e. that the sum of probabilities on all pairwise exclusive possibilities must add up to one. Instead belief theory gives observers the ability to assign so-called belief mass to any subset of the frame of discernment, i.e. non-exclusive possibilities including the whole frame itself. The main advantage of this approach is that ignorance, i.e. the lack of information, can be explicitly expressed e.g. by assigning belief mass to the whole frame. Shafer's book [1] describes many aspects of belief theory, but the two main elements are 1) a flexible way of expressing beliefs, and 2) a method for combining beliefs, commonly known as Dempster's Rule.

There are well known examples where Dempster's rule produces counterintuitive results, especially in case of strong conflict between the two argument beliefs. Mo-tivated by this observation, numerous authors have proposed alternative methods for fusing beliefs [2–9]. These rules express different behaviours with respect to the beliefs to be combined, but have been proposed with the same basic purpose in mind: to combine two beliefs into a single belief that reflects the two possibly conflicting beliefs in a fair and equal way. The considerable effort that is being invested into finding a satisfactory solution to the problem of conflict management indicates that is is considered to be of major importance in the belief theory community. However, this can also be considered to be the tragedy of belief theory for two reasons. Firstly, instead of advancing belief theory, researchers have been trapped in the search for a solution to the same problem for 20 years. Secondly, this controversy has given belief theory a bad taste despite its obvious advantages in uncertainty management.

In this article we present an alternative to Dempster's rule called the cumulative rule of combining beliefs. To be correct, these two rules are applicable to two different types of situations, although the difference between the two types is can be subtle and difficult to identify. The term *cumulative rule* was explicitly chosen in order to have a descriptive terminology for separating between the two types of situations. The conjunctive rule of combination is applicable to situations where beliefs are combined by accumulation. This is different from Dempster's rule, where beliefs are combined by normalised conjunction. This difference will be illustrated by examples below.

The cumulative rule has the limitation that it can only be directly applied to beliefs with a particular structure, which we call Dirichlet beliefs. As the name Dirichlet beliefs indicates, these belief structures are equivalent to Dirichlet probability density functions. The cumulative rule itself is then simply equivalent to the combination of Dirichlet probability density functions. In this way, belief fusion in the form of the cumulative rule is firmly based on classical statistical theory.

## 2  Belief Theory

In this section several concepts of the Dempster-Shafer theory of evidence [1] are recalled in order to introduce notations used throughout the article. Let $\Theta = \{\theta_i, i = 1, \cdots, k\}$ denote a finite set of exhaustive and exclusive possible values for a state variable of interest. The frame of discernment can for example be the set of six possible outcomes of throwing a dice, and the (unknown) outcome of a particular instance of throwing the dice becomes the state variable. A bba (basic belief assignment [1]), denoted by $m$. is defined as a belief distribution function from $2^\Theta$ to

---

[1] Called *basic probability assignment* in [1], and *Belief Mass Assignment* (BMA) in Jos2001-IJUFKS.

[0, 1] satisfying:
$$m(\emptyset) = 0 \quad \text{and} \quad \sum_{x \subseteq \Theta} m(x) = 1 . \tag{1}$$

Values of a bba are called *belief masses*. Each subset $x \subseteq \Theta$ such that $m(x) > 0$ is called a focal element of $m$.

Fig.1 below illustrates a frame of discernment $\Theta = \{x_1, x_2, x_3, x_4\}$, where shaded ellipses or circles around subsets of $\Theta$ indicates belief masses on those subsets. It can be seen that the focal elements in this general example are $\Theta$, $x_1$, $x_2$, $x_4$, $x_5$ and $x_6$. This simplistic visualisation only shows the focal elements without showing the amount of belief mass assigned to those focal elements.

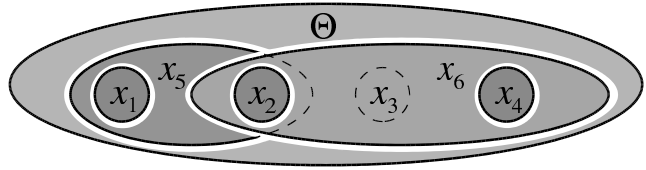

Fig. 1. A general bba on a frame of discernment containing four singletons

A bba $m$ can be equivalently represented by a non additive measure: a belief function Bel: $2^\Theta \to [0, 1]$, defined as

$$\text{Bel}(x) \triangleq \sum_{\emptyset \neq y \subseteq x} m(y) \quad \forall\, x \subseteq \Theta . \tag{2}$$

The quantity $\text{Bel}(x)$ can be interpreted as a measure of one's total belief committed to the hypothesis that $x$ is true. Note that functions $m$ and Bel are in one-to-one correspondence [1] and can be seen as two facets of the same piece of information.

A few special classes of bba can be mentioned. A vacuous bba has $m(\Theta) = 1$, i.e. no belief mass committed to any proper subset of $\Theta$. A *Bayesian* bba is when all the focal elements are singletons, i.e. one-element subsets of $\Theta$. If all the focal elements are nestable (i.e. linearly ordered by inclusion) then we speak about *consonant* bba. A *dogmatic* bba is defined by Smets as a bba for which $m(\Theta) = 0$. Let us note, that trivially, every Bayesian bba is dogmatic.

Fig.2 below illustrates a Bayesian bba, where shaded circles around singletons of $\Theta$ indicates belief masses. The focal elements in this example are the singletons $x_1$, $x_2$ and $x_4$.

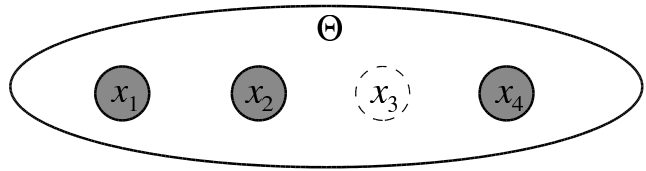

Fig. 2. A Bayesian bba characterised by belief masses on singletons only In addition, we define here the Dirichlet bba and, and its cluster variant. Dirichlet bbas are characterised by allowing only singletons and Θ as focal elements.

**Definition 1 (Dirichlet bba)**  *A bba where the only focal elements are Θ and/or singletons of Θ, is called a Dirichlet belief distribution function.*

Fig.3 below illustrates a Dirichlet bba, where the shaded circles around singletons and the shaded ellipse around Θ represent belief masses on those subsets. The focal elements in this example are Θ, $x_1$, $x_2$ and $x_4$.

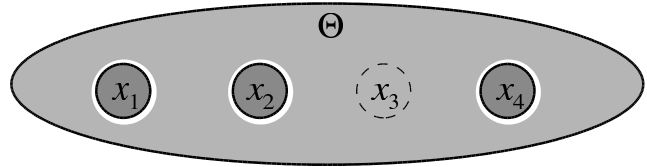

Fig. 3. A Dirichlet bba, characterised by belief masses on singletons and on Θ

It can be noted that Bayesian bbas are a special case of Dirichlet bbas.

Cluster Dirichlet bbas, defined next, are characterised by having focal elements that are mutually disjoint, except Θ itself.

**Definition 2 (Cluster Dirichlet bba)**  *A bba where the only focal elements are Θ and/or mutually exclusive subsets of Θ (singletons or clusters of singletons), is called a cluster Dirichlet belief distribution function.*

Fig.4 below illustrates a cluster Dirichlet bba, where the shaded circles and ellipses around subsets of Θ represent belief masses on those subsets. The focal elements in this example are Θ, $x_4$ and $x_5$.

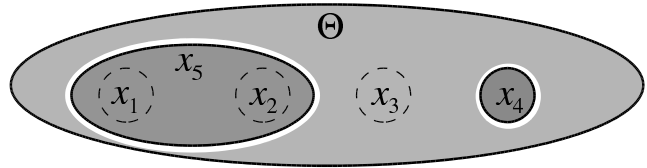

Fig. 4. A cluster Dirichlet bba, with belief masses on disjoint subsets as well as and on Θ

The probability transformation [11], also known as the pignistic transformation [12,13], produces a probability expectation value, denoted by $\wp(x)$, defined as:

$$\wp(x) = \sum_{y \in \Theta} m(y) \frac{|x \cap y|}{|y|}, \quad y \in 2^{\Theta} . \tag{3}$$

For (cluster) Dirichlet belief distribution functions, the expression for the probability expectation value is particularly simple:

$$\wp(\theta_i) = m(\theta_i) + m(\Theta)/k , \text{ when } m \text{ is a (cluster) Dirichlet bba.} \tag{4}$$

# 3 The Dirichlet Multinomial Model

The cumulative rule of combination, to be described in detail in the following sections, is firmly rooted in the classical Bayesian inference theory, and is equivalent to the combination of multinomial observations. For self-containment, we briefly outline the Dirichlet multinomial model below, and refer to [14] for more details.

## 3.1 The Dirichlet Distribution

We are interested in knowing the probability distribution over the disjoint elements of a frame of discernment. In case of a binary frame of discernment, it is determined by the beta distribution. In the general case it is determined by the Dirichlet distribution, which describes the probability distribution over a $k$-component random variable $p(\theta_i)$, $i = 1 \ldots k$ with sample space $[0,1]^k$, subject to $\sum_{i=1}^{k} p(\theta_i) = 1$.

Note that for any sample from a Dirichlet random variable, it is sufficient to determine values for $p(\theta_i)$ for any $k-1$ elements $i$ of $\{1, \ldots, k\}$, as this uniquely determines the value of the other variable.

The Dirichlet distribution captures a sequence of observations of the $k$ possible outcomes with $k$ positive real parameters $\alpha(\theta_i)$, $i = 1 \ldots k$, each corresponding to one of the possible outcomes. In order to have a compact notation we define a vector $\vec{p} = \{p(\theta_i) \mid 1 \leq i \leq k\}$ to denote the $k$-component random probability variable, and a vector $\vec{\alpha} = \{\alpha_i \mid 1 \leq i \leq k\}$ to denote the $k$-component random observation variable $[\alpha(\theta_i)]_{i=1}^{k}$.

The Dirichlet probability density function is then given by

$$f(\vec{p} \mid \vec{\alpha}) = \frac{\Gamma\left(\sum_{i=1}^{k} \alpha(\theta_i)\right)}{\prod_{i=1}^{k} \Gamma(\alpha(\theta_i))} \prod_{i=1}^{k} p(\theta_i)^{\alpha(\theta_i)-1} , \qquad (5)$$

where $p(\theta_i), \ldots, p(\theta_k) \geq 0$, $\sum_{i=1}^{k} p(\theta_i) = 1$ and $\alpha(\theta_1), \ldots, \alpha(\theta_k) > 0$.

The probability expectation value of any of the $k$ random variables is defined as:

$$\mathrm{E}(p(\theta_i) \mid \vec{\alpha}) = \frac{\alpha(\theta_i)}{\sum_{i=1}^{k} \alpha(\theta_i)} . \qquad (6)$$

Because of the additivity requirement $\sum_{i=1}^{k} p(\theta_i) = 1$, the Dirichlet distribution has only $k-1$ degrees of freedom. This means that knowing $k-1$ probability variables and their density uniquely determines the last probability variable and its density.

*3.2 A Priori Distribution for $k$ Alternatives*

Now, we come to the question of an *a priori* density function for the probabilities of $k$ exhaustive and mutually exclusive alternatives (*e.g.* $k$ different colours of balls in an urn). Let $p(\theta_i)$ denote the random variable describing the probability of a random sample (*e.g.* drawing a ball of a particular colour) yielding alternative $i$. Since $p(\theta_i)$ describes a probability, then the sample space for $(p(\theta_i))_{i=1}^{k}$ is $[0,1]^k$. Since the alternatives are exhaustive and mutually exclusive, then

$$\sum_{i=1}^{k} p(\theta_i) = 1. \tag{7}$$

Generalising the case of 2 alternatives, where the probability density function is commonly called the Beta distribution, we will take an *a priori* Dirichlet distribution. Since there is no reason to assume a preference for any alternative over any other alternative, then the parameters will be taken to be equal, with the result that $\mathrm{E}(p(\theta_i)) = \frac{1}{k}$ for all $i$. In the case of 2 alternatives, a uniform distribution must be assumed (*i.e.* Beta$(1,1)$). The question arises as to whether this fact can be used to determine the common value of the parameters in the case of $k$ alternatives on the grounds of consistency. It can be argued that such a determination is possible, and that the common value of the parameters is $\frac{2}{k}$. The argument goes as follows.

For integers $v$ and $w$, take a set of $vw$ exhaustive mutually exclusive alternatives, and a partition of the set into $v$ classes, each with $w$ elements. The *a priori* distribution for the probabilities of the $vw$ alternatives is a Dirichlet distribution with the common value of the parameters being given by $\alpha_{vw}$ (here, $\alpha_k$ denotes the common value of the parameters in the case where there are $k$ alternatives). It follows that for the partition, the distribution for the probabilities of the $v$ alternative classes is a Dirichlet distribution with a common value for the parameters, equal to $w\alpha_{vw}$. Since the $v$ classes are exhaustive mutually exclusive alternatives in their own right, with no reason for preference for any over the others, then the distribution for the probabilities should have a common value of the parameters equal to $\alpha_v$, and so consistency requires that $\alpha_v = w\alpha_{vw}$. Since

$$v\alpha_v = vw\alpha_{vw} = w\alpha_w$$

for all positive integers $v$, $w$, then $\alpha_w = \frac{C}{w}$ for some constant $C$. The crucial step is now to define the value of this *a priori* constant. It is normally required that the *a priori* distribution in case of 2 alternatives is uniform, which means that $\alpha_2 = 1$. Then necessarily $C = 2$, and the common value of the parameters in the case of $k$

alternatives is:

$$\alpha_k = \frac{C}{k} = \frac{2}{k} \ . \tag{8}$$

Should one assume an *a priori* distribution over binary state spaces different from uniform, the constant, and also the common value would be different. The constant $C$ will always be equal to the size of the state space over which a uniform distribution is assumed. The constant $C = 2$ emerges precisely because we assume a uniform distribution over binary state spaces. This means that the *a priori* distribution over state spaces larger than binary will not be uniform. A justification for the choice $C = 2$ is given below.

The state space size provides *a priori* information about the *base rate* of an arbitrary event out of the $k$ possible events. We define the base rate $a_k$ for any of the $k$ singleton events of a state space if size $k$ as:

$$a_k = \frac{1}{k} \ . \tag{9}$$

In case no other evidence is available, the base rate alone determines the probability distribution of the events. For example in the binary case, the *a priori* probability of any of the two possible outcomes is $\frac{1}{2}$, and the probability density function is the uniform $\text{Beta}(1,1)$. As more evidence becomes available, the influence of the base rate is reduced, until the evidence alone determines the probability distribution of the events. It is thus possible to separate between the *a priori* base rate denoted by $a_k$ and the *a posteriori* evidence over the possible events denoted as a vector $\vec{r}$. The total evidence $\alpha(\theta_i)$ for each singleton event $\theta_i$ can then be expressed as:

$$\alpha(\theta_i) = r(\theta_i) + Ca_k \tag{10}$$

In order to distinguish between the *a priori* and the *a posteriori* evidence, we introduce the augmented notation for Dirichlet distribution over a set of $k$ singletons as:

$$f(\vec{p} \mid \vec{r}, a_k) = \frac{\Gamma\left(\sum_{i=1}^{k}\left(r(\theta_i) + a_k C\right)\right)}{\prod_{i=1}^{k} \Gamma\left(r(\theta_i) + a_k C\right)} \prod_{i=1}^{k} p(\theta_i)^{r(\theta_i) + a_k C - 1} \ . \tag{11}$$

This is useful, because it allows the determination of the probability distribution without, or only with little evidence. Given the augmented Dirichlet distribution of Eq.(11), the probability expectation of any of the $k$ random variables can now be written as:

$$\mathrm{E}(p(\theta_i) \mid \vec{r}, a_k) = \frac{r(\theta_i) + a_k C}{C + \sum_{i=1}^{k} r(\theta_i)} \ . \tag{12}$$

As a justification for our choice of the *a priori* constant value $C = 2$, it can be noted that it would be unnatural to require a uniform distribution over arbitrary large state spaces because it would make the sensitivity to new evidence arbitrary small. For example, when requiring a uniform *a priori* distribution over a state space of cardinality 100 (thereby forcing $C = 100$), the derived probability expectation of an event that has been observed 100 times would still only be about $\frac{1}{2}$, which would seem totally counterintuitive. By assuming a uniform distribution in the binary case, the derived probability expectation of the same event after 100 identical observations would be close to 1, as intuition would dictate.

### 3.3 Visualising Dirichlet Distributions

Visualising Dirichlet distributions is challenging because it is a density function over $k-1$ dimensions, where $k$ is the state space size. For this reason, Dirichlet distributions over ternary state spaces are the largest that can be easily visualised.

With $k = 3$, the probability distribution has 2 degrees of freedom, and the equation $p(\theta_1) + p(\theta_2) + p(\theta_3) = 1$ defines a triangular plane as illustrated in Fig.5.

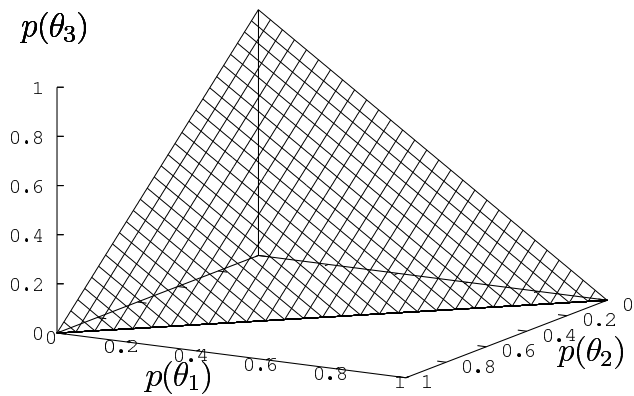

Fig. 5. Triangular plane

In order to visualise probability density over the triangular plane, it is convenient to lay the triangular plane horizontally in the $x$-$y$ plane, and visualise the density dimension along the $z$-axis.

Let is consider the example of an urn containing balls of the three different colours: red black and yellow (i.e. $k = 3$). Let us first assume that no other information than the size of the state space is available, meaning that $r(\text{red}) = r(\text{black}) = r(\text{yellow}) = 0$. Then Eq.(9) and Eq.(12) dictate that the expected probability of picking a ball of any specific colour is the *a priori* base rate probability, which is $\frac{1}{3}$. The *a priori* Dirichlet density function is illustrated in Fig.6.

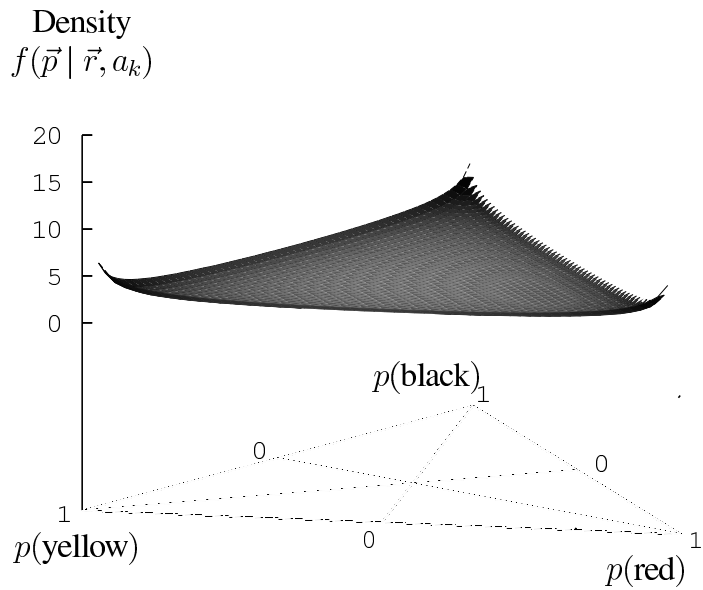

Fig. 6. Prior Dirichlet distribution in case of urn with balls of 3 different colours

Let us now assume that an observer has picked (with return) 6 red, 1 black and 1 yellow ball, i.e. $r(\text{red}) = 6$, $r(\text{black}) = 1$, $r(\text{yellow}) = 1$, then the *a posteriori* expected probability of picking a red ball can be computed as $\mathrm{E}(p(\text{red})) = \frac{2}{3}$. The *a posteriori* Dirichlet density function is illustrated in Fig.7.

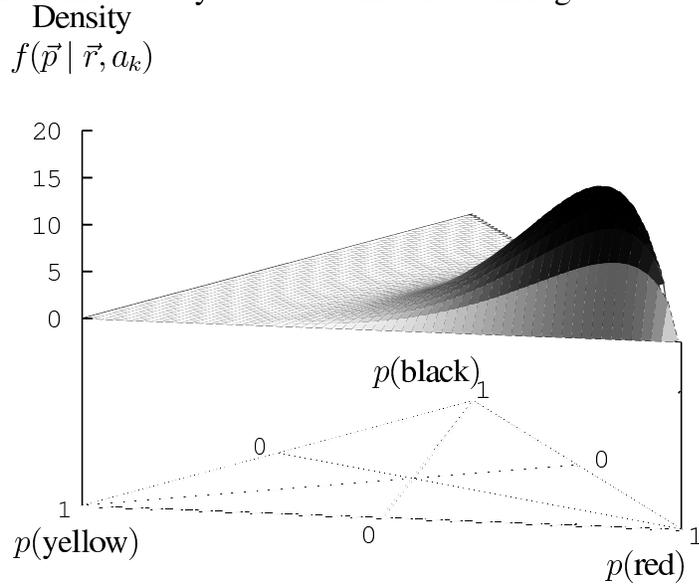

Fig. 7. *A posteriori* Dirichlet distribution after picking 6 red, 1 black and 1 yellow ball

### 3.4 Cluster Dirichlet Distribution

We now ask what happens if instead of considering the probability distribution functions $[p(\theta_i)]_{i=1}^{k}$ over singletons of $\Theta$, we consider the probability distribution functions over $l$ mutually disjoint proper subsets $x_j$ of $\Theta$ so that $\Theta = \bigcup_{j=1}^{l} x_j$. Each subset $x_j$ can contain one or several singletons, but no subset can contain all singletons. Let $\Theta_X = \{x_j \mid j = 1 \ldots l\}$ be such a nontrivial coarsening of $\Theta$ where $1 < l < k$. We do not consider the trivial case $l = k$, where every subset would be a singleton, because the coarsened frame would be identical to the original frame. We also do not consider the trivial case $l = 1$, where the base rate of the only subset $x_1$ would be 1, meaning that the event $x_1$ would always happen with absolute certainty.

We are thus interested in the probability distribution functions $[p(x_j)]_{j=1}^{l}$ over the cluster subsets $x_j$ of $\Theta$, where $p(x_j) = \sum_{\theta_i \in x_j} p(\theta_i)$ and the evidence parameters $\alpha(x_j) = \sum_{\theta_i \in x_j} \alpha(\theta_i)$. The distribution is given by:

**Theorem 1 (Cluster Dirichlet Distributions)**

$$\text{If } [p(\theta_i)]_{i=1}^{k} \sim \text{Dirichlet}(\alpha(\theta_1), ..., \alpha(\theta_k)) , \tag{13}$$

$$\text{then } [p(x_j)]_{j=1}^{l} \sim \text{Dirichlet}(\alpha(x_1), ..., \alpha(x_l)) , \tag{14}$$

*i.e. the distribution is still a Dirichlet distribution, and the parameter corresponding to a specific sum of random variables is given by the sum of the parameters corresponding to the constituent addends.*

The proof of this theorem can be found in in [15], as well as in standard textbooks.

The prior base rate for a specific cluster $x$ of possible states is simply the sum of the base rates of all singletons contained in the cluster. The cardinality $|x|$ of each subset $x \in \Theta_{\mathbb{C}}$ reflects the number of singletons that $x$ contains. The base rate of $x$ can then be expressed as:

$$a(x) = \sum_{\theta_i \in x} a(\theta_i) = |x| a_k = \frac{|x|}{k} \tag{15}$$

The total evidence $\alpha(x_j)$ for each cluster event $x_j$ can then be expressed as:

$$\alpha(x_j) = r(x_j) + a(x_j) \tag{16}$$

The augmented Dirichlet distribution notation can be expressed similarly to Eq.(11) as:

$$f(\vec{p} \mid \vec{r}, \vec{a}) = \frac{\Gamma\left(\sum_{j=1}^{l} (r(x_j) + Ca(x_j))\right)}{\prod_{j=1}^{l} \Gamma\left(r(x_j) + Ca(x_j)\right)} \prod_{j=1}^{l} p(x_j)^{r(x_j) + Ca(x_j) - 1} . \tag{17}$$

This is again useful, because it allows the determination of the probability distribution over coarsened frames of discernment. Given the augmented Dirichlet distribu-tion of Eq.(17), the probability expectation of any of the $l$ random cluster variables can now be written as:

$$\mathrm{E}(p(x_j) \mid \vec{r}, \vec{a}) = \frac{r(x_j) + Ca(x_j)}{C + \sum_{j=1}^{l} r(x_j)} \ . \tag{18}$$

We reuse the example of the urn containing red, black and yellow balls, but this time we create a binary partition of $x_1 = \{\text{red}\}$ and $x_2 = \{\text{black}, \text{yellow}\}$. The base rate of picking a red ball can be computed with Eq.(15) as $a(x_1) = \frac{1}{3}$.

Let us again assume that an observer has picked (with return) 6 red balls, and 2 black or yellow balls, i.e. $r(x_1) = 6$, $r(x_2) = 2$.

Since the state space has been reduced to binary, the Dirichlet distribution is reduced to a Beta distribution which is simple to visualise. The *a priori* and *a posteriori* density functions are illustrated in Fig.8.

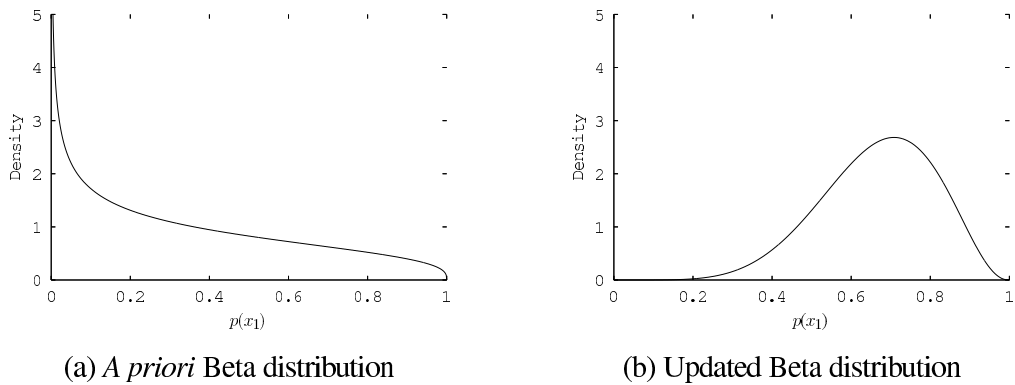

(a) *A priori* Beta distribution  (b) Updated Beta distribution

Fig. 8. *A priori* and *a posteriori* Beta distributions

The *a posteriori* expected probability of picking a red ball can be computed with Eq.(18) as $\mathrm{E}(p(x_1)) = \frac{2}{3}$, which is the same as before the coarsening, as illustrated in Sec.3.3. This shows that the coarsening does not influence the probability expectation value of specific events.

## 4 Mapping Between Dirichlet Distribution and Belief Distribution Functions

In this section we will define a bijective mapping between Dirichlet probability distributions described in Sec.3, and Dirichlet belief distribution functions described in Sec.2.

Assume a frame of discernment $\Theta = \{\theta_i \mid i = 1, \cdots, k\}$, where each singleton represents a possible outcome of a state variable, and assume $\Theta_X = \{x_j \mid j =$ $1, \cdots, l\}$ to be a coarsening of $\Theta$, i.e. a partition into $l$ mutually exclusive elements where $k \geq l$. In the trivial case $k = l$ and $\theta_i = x_j$, meaning that $X$ is identical to, and not a coarsening of $\Theta$. Let $m$ be a Dirichlet bba on $\Theta$, and let $f(\vec{p} \mid \vec{r}, \vec{a})$ be an augmented Dirichlet probability distribution function over $\Theta$.

For the bijective mapping between $m$ and $f(\vec{p} \mid \vec{r}, \vec{a})$, we require equality between the pignistic probability values $\wp(x_j)$ derived from $m$, and the probability expectation values $\mathrm{E}(p(x_j))$ of $f(\vec{p} \mid \vec{r}, \vec{a})$. This constraint is expressed as:

For all $x_j \in \Theta_X$:

$$\wp(x_j) = \mathrm{E}(p(x_j) \mid \vec{r}, \vec{a}) \tag{19}$$
$$\Updownarrow$$
$$m(x_j) + a(x_j) m(\Theta) = \frac{\alpha(x_j)}{C + \sum_{j=1}^{l} r(x_j)} + \frac{Ca(x_j)}{C + \sum_{j=1}^{l} r(x_j)} \tag{20}$$

We also require that $m(x_j)$ be an increasing function of $\alpha(x_j)$, and that $m(\Theta)$ be a decreasing function of $\sum_{j=1}^{l} \alpha(x_j)$. In other words, the more evidence in favour of a particular outcome, the greater the belief mass on that outcome. Furthermore, the less evidence available in general, the more vacuous the bba (i.e. the greater $m(\Theta)$). These intuitive requirements together with Eq.(20) imply the following bijective mapping:

$$\text{For } m(\Theta) \neq 0: \begin{cases} m(x_j) = \frac{r(x_j)}{C + \sum_{i=1}^{l} r(x_i)} \\ m(\Theta) = \frac{C}{C + \sum_{i=1}^{l} r(x_i)} \end{cases} \Leftrightarrow \begin{cases} r(x_j) = \frac{Cm(x_j)}{m(\Theta)} \\ 1 = m(\Theta) + \sum_{i=1}^{k} m(x_i) \end{cases} \tag{21}$$

In case $m(\Theta) \longrightarrow 0$, then necessarily $\sum_{j=1}^{k} m(x_j) \longrightarrow 1$, and $\sum_{j=1}^{k} r(x_j) \longrightarrow \infty$, meaning that at least some, but not necessarily all, of the evidence parameters $r(x_j)$ are infinite. We define $\eta(x_j)$ as the the relative degree of infinity between the corresponding infinite evidence parameters $r(x_j)$ such that $\sum_{j=1}^{l} \eta(x_j) = 1$. When infinite evidence parameters exist, any finite evidence parameter $r(x_j)$ can be assumed to be zero in any practical situation because it will have $\eta(x_j) = 0$, i.e. it will carry zero weight relative to the infinite evidence parameters. This leads to the following bijective mapping:

For
$m(\Theta) = 0$ :
$$\begin{cases} m(x_j) = \eta(x_j) \\ m(\Theta) = 0 \end{cases} \Leftrightarrow \begin{cases} r(x_j) = \eta(x_j) \sum_{i=1}^{l} r(x_i) = \eta(x_j)\infty \\ 1 = \sum_{j=1}^{l} m(x_j) \end{cases} \quad (22)$$

In case $\eta(x_j) = 1$ for a particular evidence parameter $r(x_j)$, then $r(x_j) = \infty$ and all the other evidence parameters are finite. In case $\eta(x_j) = 1/l$ for all $j = 1\ldots l$, then all the evidence parameters are all equally infinite.

## 5  Deriving the Cumulative Rule of Belief Fusion

The cumulative rule is equivalent to *a posteriori* updating of Dirichlet distributions. Its derivation is based on the bijective mapping between Dirichlet distributions and Dirichlet bbas described in the previous section.

Assume a process with $k$ possible outcomes, or equivalently a frame of discernment $\Theta$ containing $k$ singletons. Assume two observers $A$ and $B$ who observe the outcomes of the process over two separate time periods. They may group observations of different outcomes into clusters, so that they in fact create a coarsened frame of discernment $\Theta_X = \{x_j \mid j = 1, \cdots, l\}$ consisting of $l$ mutually exclusive subsets. It is required that the two observers use the same partitioning. This implies a common base rate vector $\vec{a}$ where $a(x_j) = |x_j|/k$, as defined by Eq.(15).

Let the two observers' respective observations be expressed as $\vec{r}_A$ and $\vec{r}_B$. The augmented Dirichlet distributions resulting from these separate bodies of evidence can be expressed as $f(\vec{p} \mid \vec{r}_A, \vec{a})$ and $f(\vec{p} \mid \vec{r}_B, \vec{a})$

The fusion of these two bodies of evidence simply consists of vector addition of $\vec{r}_A$ and $\vec{r}_B$. Expressed in terms of Dirichlet distributions, this can be expressed as:

$$f(\vec{p} \mid \vec{r}_A, \vec{a}) \oplus f(\vec{p} \mid \vec{r}_B, \vec{a}) = f(\vec{p} \mid (\vec{r}_A + \vec{r}_B), \vec{a}) \ . \quad (23)$$

The symbol "$\diamond$" denotes the fusion of two observers $A$ and $B$ into a single imaginary observer denoted as $A \diamond B$. All the necessary elements are now in place for presenting the cumulative rule for belief fusion.

**Theorem 2 (Cumulative Rule)**
*Let $m_A$ and $_B$ be Dirichlet bbas respectively held by agents $A$ and $B$ over the same*

*partitioning* $\Theta_X = \{x_j \mid j = 1, \cdots, l\}$ *of a frame of discernment* $\Theta$, *meaning that* $m_A$ *and* $_B$ *are cluster Dirichlet bbas over* $\Theta$. *Let* $m_{A \diamond B}$ *be the bba such that:*

*Case I:* For $m_A(\Theta) \neq 0 \ \vee \ m_B(\Theta) \neq 0$:

$$\begin{cases} m_{A \diamond B}(x_j) = \dfrac{\left(\dfrac{m_A(x_j)}{m_A(\Theta)} + \dfrac{m_B(x_j)}{m_B(\Theta)}\right)}{1 + \sum_{j=1}^{k}\left(\dfrac{m_A(x_j)}{m_A(\Theta)} + \dfrac{m_B(x_j)}{m_B(\Theta)}\right)} \\ \\ m_{A \diamond B}(\Theta) = \dfrac{1}{1 + \sum_{j=1}^{k}\left(\dfrac{m_A(x_j)}{m_A(\Theta)} + \dfrac{m_B(x_j)}{m_B(\Theta)}\right)} \end{cases} \quad (24)$$

*Case II:* For $m_A(\Theta) = 0 \ \wedge \ m_B(\Theta) = 0$:

$$\begin{cases} m_{A \diamond B}(x_j) = \gamma_A \, m_A(x_j) + \gamma_B m_B(x_j) \\ \\ m_{A \diamond B}(\Theta) = 0 \end{cases} \quad (25)$$

$$\text{where} \quad \gamma_A = \lim_{\substack{m_A(\Theta) \to 0 \\ m_B(\Theta) \to 0}} \frac{m_B(\Theta)}{m_A(\Theta) + m_B(\Theta)}$$

$$\text{and} \quad \gamma_B = \lim_{\substack{m_A(\Theta) \to 0 \\ m_B(\Theta) \to 0}} \frac{m_A(\Theta)}{m_A(\Theta) + m_B(\Theta)}$$

*Then* $m_{A \diamond B}$ *is called the cumulatively fused bba of* $m_A$ *and* $m_B$, *representing an imaginary agent* $[A \diamond B]$*'s bba, as if that agent represented both A and B. By using the symbol '$\oplus$' to designate this belief operator, we define* $m_{A \diamond B} \equiv m_A \oplus m_B$.

The proof below provides details about how the expression for the cumulative rule can be derived.

**Proof 1** *Let $m_A$ and $m_B$ be (cluster) Dirichlet bbas (with equal partitioning). The mapping from (cluster) Dirichlet bbas to augmented Dirichlet distributions is done according to the right sides of Eq.(21) and Eq.(22), expressed as:*

$$m_A \longmapsto f(\vec{p} \mid \vec{r}_A, \vec{a})$$
$$m_B \longmapsto f(\vec{p} \mid \vec{r}_B, \vec{a})$$
(26)

*These augmented Dirichlet distributions can now be fused according to Eq.(23), expressed as:*

$$f(\vec{p} \mid \vec{r}_A, \vec{a}) \oplus f(\vec{p} \mid \vec{r}_B, \vec{a}) = f(\vec{p} \mid (\vec{r}_A + \vec{r}_B), \vec{a}) \qquad (27)$$

*Finally, the result of Eq.(27) is mapped back to a (cluster) Dirichlet bba again using the left sides of Eq.(21) and Eq.(22). This can be written as:*

$$f(\vec{p} \mid (\vec{r}_A + \vec{r}_B), \vec{a}) \longmapsto m_{A \diamond B} \qquad (28)$$

*By inserting the full expressions for the parameters in Eqs.(26), (27) and (28), the expressions of Eqs.(24) and (25) in Theorem 2 emerge.*

□

It can be verified that the cumulative rule is commutative, associative and non-idempotent. In Case II of Theorem 2 (Bayesian bbas, which can also be described as dogmatic Dirichlet bbas), the associativity depends on the preservation of relative weights of intermediate results, which requires the additional weight variable $\gamma$. In this case, the cumulative rule is equivalent to the weighted average of probabilities.

It is interesting to notice that the expression for the cumulative rule is independent of the *a priori* constant $C$. That means that the choice of a uniform Dirichlet distribution in the binary case in fact only influences the mapping between Dirichlet distributions and Dirichlet bbas, not the cumulative rule itself. This shows that the cumulative rule is firmly based on classical statistical analysis, and not dependent on arbitrary and ad hoc choices.

The cumulative rule represents a generalisation of the consensus operator [5,10] which emerges directly from Theorem 2 by assuming a binary state space.

## 6 Examples

In this section, we will illustrate by examples the results of applying the cumulative rule of fusing beliefs.

### 6.1 Zadeh's Example

This well known example was put forward by Zadeh [16] to show that Dempster's rule produces counterintuitive results when applied to particular situations. We will show that the cumulative rule produces a result which is well in line with intuition.

For comparison, we also include the results of Dempster's rule. The definitions of Conjunctive Rule and of Dempster's Rule are given below:

**Definition 3 (The Conjunctive Rule) .**

$$[m_A \odot m_B](x) = \sum_{y \cap z = x} m_A(y) m_B(z) \quad \forall\, x \subseteq \Theta. \tag{29}$$

This rule is referred to as the conjunctive rule of combination, or the non-normalised Dempster's rule. If necessary, the normality assumption $m(\emptyset) = 0$ can be recovered by dividing each mass by a normalisation coefficient. The resulting operator known as Dempster's rule is defined as:

**Definition 4 (Dempster's Rule) .**

$$[m_A \odot m_B](x) = \frac{[m_A \odot m_B](x)}{1 - [m_A \odot m_B](\emptyset)} \quad \forall\, x \subseteq \Theta,\ x \neq \emptyset \tag{30}$$

The use of Dempster's rule is possible only if $m_A$ and $m_B$ are not totally conflicting, i.e., if there exist two focal elements $y$ and $z$ of $m_A$ and $m_B$ satisfying $y \cap z \neq \emptyset$.

Let us now return to Zadeh's example. Suppose that we have a murder case with three suspects; Peter, Paul and Mary, and two witnesses $W_A$ and $W_B$ who give highly conflicting testimonies. The example assumes that the most reasonable conclusion about the likely murderer can be obtained by fusing the beliefs expressed by the two witnesses using Dempster's rule. Table 1 gives the witnesses' belief masses in Zadeh's example and the resulting belief masses after applying Dempster's rule.

Since this example involved Bayesian bbas (dogmatic Dirichlet bbas), Case II of Theorem 2 applies.

|  | Witness A | Witness B | Cumulative Rule | Dempster's Rule |
|---|---|---|---|---|
| $m(\text{Peter}) =$ | 0.99 | 0.00 | 0.495 | 0.00 |
| $m(\text{Paul}) =$ | 0.01 | 0.01 | 0.010 | 1.00 |
| $m(\text{Mary}) =$ | 0.00 | 0.99 | 0.495 | 0.00 |

Table 1
Zadeh's example of belief fusion with the Cumulative Rule and with Dempster's Rule

In case of Bayesian bbas, the cumulative rule thus is equivalent to weighted average of probabilities. The next example will illustrate the results of fusing non-dogmatic Dirichlet bbas.

## 6.2 Fusing Non-Conflicting Equal Sensor Outputs

Assume that a manufacturing process can produce Correct and Faulty parts, and that two sensors observe the likelihood with which Correct and Faulty parts are produced. The frame of discernment has two elements: $\Theta = \{\text{Correct}, \text{Faulty}\}$ and the powerset three elements: $2^\Theta = \{\text{Correct}, \text{Faulty}, \Theta\}$. For the purpose of independence it can be assumed that the sensors observe the manufacturing process at different time periods.

Assume that the two sensors produce equal and non-conflicting beliefs as given in the Table 2 below:

|  |  | Sensor $A$ | Sensor $B$ | Cumulative rule | Dempster's rule |
|---|---|---|---|---|---|
| $m(\text{Correct})$ | = | 0.99 | 0.99 | 0.994975 | 0.9999 |
| $m(\text{Faulty})$ | = | 0.00 | 0.00 | 0.00 | 0.00 |
| $m(\Theta)$ | = | 0.01 | 0.01 | 0.005025 | 0.0001 |

Table 2
Results of combining equal non-conflicting beliefs from two sensors

Applying the cumulative rule and Dempster's rule to these beliefs results in a reduction in uncertainty, and a convergence towards the largest belief mass of the sensor outputs.

In case of the cumulative rule, the input beliefs are equivalent to each sensor having observed 198 Correct parts ($r(\text{Correct}) = 198$), and no Faulty parts ($r(\text{Faulty}) = 0$), with the uncertainty computed as $m_{A \diamond B}(\Theta) = 2/(r(\text{Correct}) + r(\text{Faulty}) + 2) = 0.01$. The output beliefs is equivalent to the observation of $2r(\text{Correct}) = 396$ Correct parts, with the uncertainty computed as $m(\Theta) = 2/(2 \cdot 396 + 2) = 0.005025$. The output uncertainty of the cumulative rule is thus halved, which is what one would expect in the case of cumulative fusion.

In case of Dempster's rule, the uncertainty is reduced to the product of the input uncertainties computed as:

$$m_{A \cdot B}(\Theta) = m_A(\Theta)\, m_B(\Theta) = 0.0001 \tag{31}$$

i.e. by a factor of $1/100$ which represents a very fast convergence. This rate is too fast to be the result of cumulative belief fusion. When fusing two equal sensor outputs, one should expect the uncertainty to be reduced by $1/2$, because the double amount of observations have been made.

The difference in convergence i.e. the rate of uncertainty reduction, is significant,

and clearly illustrates that Dempster's rule in fact is not applicable to this situation, even in the case of non-conflicting beliefs. This is because the example describes a cumulative situation, and that it would be meaningless to model it with a conjunctive fusion rule.

# 7 Discussion and Conclusion

The cumulative rule of belief fusion allows the modelling of many practical situations with belief theory, that previously could not be modelled, or were incorrectly modelled. It is worth noticing that the cumulative rule and Dempster's rule apply to different types of belief fusion, and that it strictly speaking is meaningless to compare their performance in the same examples. The notion of cumulative belief fusion as opposed to conjunctive belief fusion has therefore been introduced in order to make this distinction explicit.

There is however considerable confusion regarding the applicability of Dempster's rule, which e.g. is illustrated by applying Dempster's rule to the court case situation in Zadeh's example. Often the problem is to identify which model best fits a particular situation. Although the court case of Zadeh's example is not as clear as in the case of the two sensors, it seems more natural to model it with the cumulative rule rather than Dempster's rule.

The mapping between beliefs and probability distribution functions bring belief theory and statistical theory closer together. This is important in order to make belief theory more practical and easier to interpret, and will hopefully contribute to removing the bad taste that belief theory has had in the main stream statistics and probability community.